\documentclass[conference]{IEEEtran}
\IEEEoverridecommandlockouts

\usepackage{cite}
\usepackage{amsmath,amssymb,amsfonts}
\usepackage{algorithmic}
\usepackage{graphicx}
\usepackage{textcomp}
\usepackage{xcolor}
\usepackage{tabularx}
\usepackage{multirow}
\usepackage{multicol}
\usepackage{booktabs}
\usepackage{colortbl}
\usepackage[colorlinks=true,urlcolor=magenta,linkcolor=blue]{hyperref}
\usepackage{enumerate}
\usepackage{enumitem}
\usepackage{orcidlink}

\def\BibTeX{{\rm B\kern-.05em{\sc i\kern-.025em b}\kern-.08em
    T\kern-.1667em\lower.7ex\hbox{E}\kern-.125emX}}

\begin{document}

\title{MicarVLMoE: A Modern Gated Cross-Aligned Vision-Language Mixture of Experts Model for Medical Image Captioning and Report Generation}

\author{
\centering
\IEEEauthorblockN{
Amaan Izhar\orcidlink{0000-0001-6394-0794}\textsuperscript{1},
Nurul Japar\orcidlink{0000-0002-3054-1874}\textsuperscript{1*},
Norisma Idris\orcidlink{0000-0002-8006-7496}\textsuperscript{1},
Ting Dang\orcidlink{0000-0003-3806-1493}\textsuperscript{2}}
\IEEEauthorblockA{\textsuperscript{1}\textit{Faculty of Computer Science and Information Technology, Universiti Malaya, Kuala Lumpur, Malaysia} \\
\textsuperscript{2}\textit{School of Computing and Information Systems, The University of Melbourne, Melbourne, Australia} \\
\textsuperscript{*}Corresponding Author (nuruljapar@um.edu.my)
}
}

\maketitle

\begin{abstract}
Medical image reporting (MIR) aims to generate structured clinical descriptions from radiological images. Existing methods struggle with fine-grained feature extraction, multimodal alignment, and generalization across diverse imaging types, often relying on vanilla transformers and focusing primarily on chest X-rays. We propose MicarVLMoE, a vision-language mixture-of-experts model with gated cross-aligned fusion, designed to address these limitations. Our architecture includes: (i) a multiscale vision encoder (MSVE) for capturing anatomical details at varying resolutions, (ii) a multihead dual-branch latent attention (MDLA) module for vision-language alignment through latent bottleneck representations, and (iii) a modulated mixture-of-experts (MoE) decoder for adaptive expert specialization. We extend MIR to CT scans, retinal imaging, MRI scans, and gross pathology images, reporting state-of-the-art results on COVCTR, MMR, PGROSS, and ROCO datasets. Extensive experiments and ablations confirm improved clinical accuracy, cross-modal alignment, and model interpretability. Code is available at \url{https://github.com/AI-14/micar-vl-moe}.
\end{abstract}


\section{Introduction}
Medical image reporting (MIR) automates the generation of structured textual descriptions from radiological images, supporting disease diagnosis, monitoring, and clinical decision-making \cite{alfarghaly2021automated, yu2024long}. It comprises two key tasks: image captioning for concise visual summaries, and report generation for detailed clinical narratives. Manual reporting is time-intensive and variable \cite{izhar2025engaging, liu2019clinically}, prompting the use of deep learning and vision-language models for more consistent automation \cite{aksoy2023radiology, xu2023vision}.

However, current MIR models face key limitations: (i) insufficient extraction of fine-grained anatomical features, (ii) suboptimal multimodal fusion and alignment, and (iii) limited evaluation beyond chest X-rays. Most methods rely on vanilla transformers -- with dense feedforward layers that uniformly aggregate features and overlook local anatomical cues, post-normalization that hinders gradient flow and stable feature learning, single-scale encoders that capture only high-level features, and standard self-attention lacking hierarchical representation -- which collectively limit precise alignment \cite{vaswani2017attention, chen2020generating}. Basic fusion strategies further hinder anatomical sensitivity and distinctions \cite{kim2021vilt, wang2023image}. These architectural bottlenecks motivate a redesign and rethinking of the MIR problem through new architectural paradigms.

To address these issues, we propose MicarVLMoE, a modern MIR framework combining: (i) a multiscale vision encoder for high-low resolution feature extraction, (ii) a dual-branch latent attention module for cross-modal alignment in latent space, and (iii) a modulated mixture-of-experts decoder for adaptive specialization. Unlike prior works, we expand beyond chest X-rays by training MicarVLMoE individually on diverse imaging modalities -- CT lungs scans, retinal Fundus/OCT/FFA imaging, gross pathology, MRI, and other 2D radiology datasets -- to assess its adaptability to varied report generation tasks. We do not perform joint training across modalities due to inherent domain differences, but instead examine generalization capabilities across individually trained models. Therefore, to guide our research, we formulate the following research questions:

\noindent
\textbf{RQ1:} How can multiscale vision encoding, latent attention, and MoE enhance fine-grained feature extraction and multimodal alignment in medical image reporting?

\noindent
\textbf{RQ2:} How versatile is MicarVLMoE in generating clinically accurate reports across different imaging modalities while maintaining clinical accuracy?

\begin{figure*}[!pbth]
    \centering
    \includegraphics[width=\linewidth, height=3in]{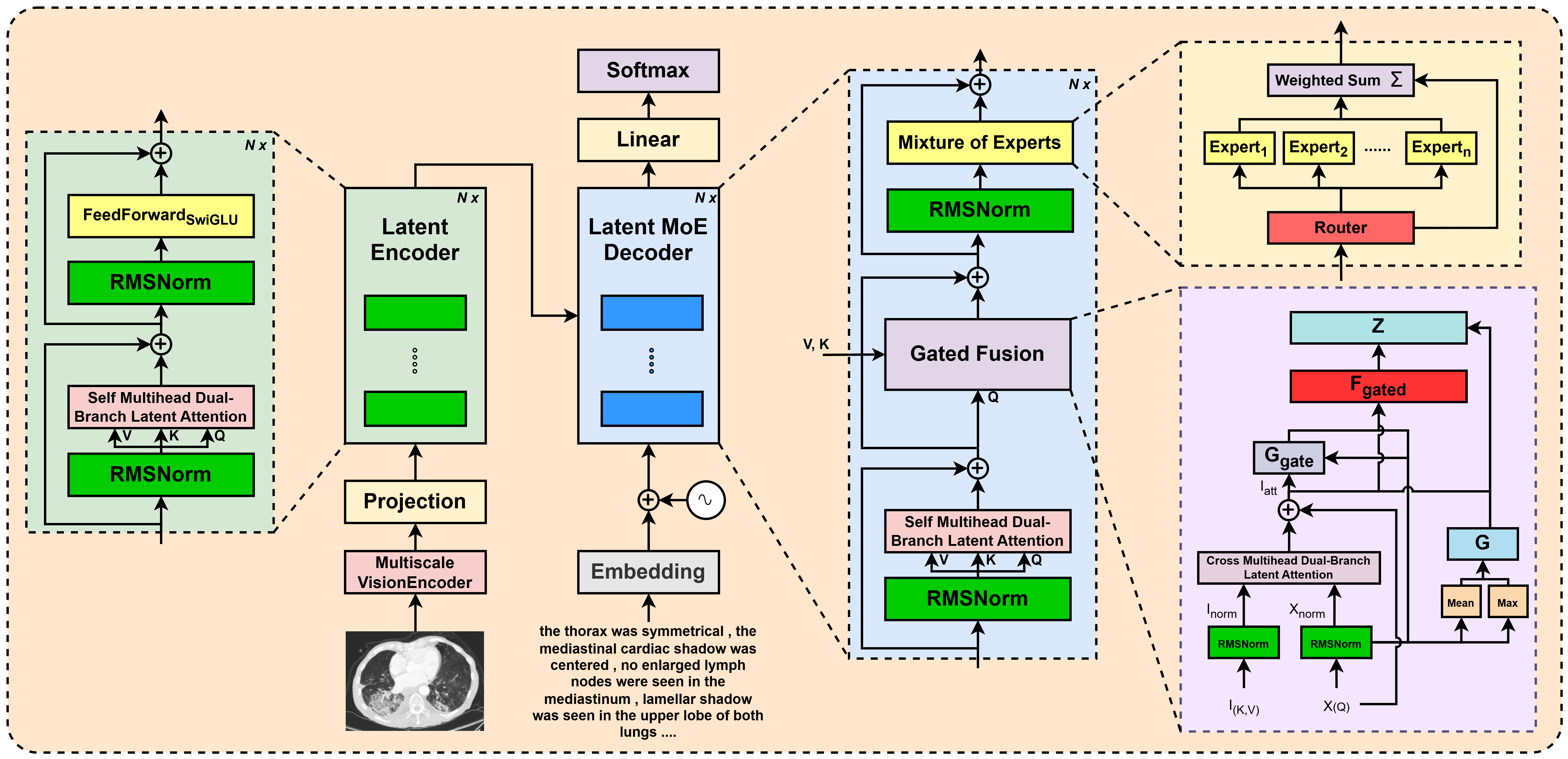}
    \caption{An overview of the proposed MicarVLMoE model highlighting the overall architecture for medical image reporting with core modules encompassing -- multiscale vision encoder, latent encoder, and latent MoE decoder with gated fusion.}
    \label{fig:arch}
\end{figure*}

Our contributions are summarized as follows:
\begin{itemize}
    \item We introduce a mixture-of-experts-based multimodal learning into MIR, leveraging sparse expert routing for efficient and scalable report generation capturing fine-grained report features via expert utilization.
    \item We enhance multimodal alignment by introducing multihead dual-branch latent attention and gated fusion modules, significantly improving contextual understanding between medical images and textual descriptions.
    \item We expand MIR evaluations beyond chest X-rays, providing new baselines across CT, retinal imaging, MRI, and gross pathology image reporting datasets.
    \item We demonstrate state-of-the-art performance across multiple datasets -- COVCTR, MMR, PGROSS, and ROCO -- improving medical single-sentence captioning and multi-sentence report generation accuracy.
    \item We conduct rigorous ablation studies and analyses, providing in-depth insights into the architectural impact of our modules -- ensuring interpretability in MIR.
\end{itemize}

\section{Related Work}
\subsection{Vision Language Models}
Transformer-based architectures have become standard in MIR, leveraging vision-language models to generate radiology reports from multimodal inputs. Several models adopt vanilla transformer backbones which act as static feature aggregators \cite{chen2020generating, xu2023vision, aksoy2023radiology, yu2024long}. These works utilize vanilla transformer architecture with post-normalization, dense feedforward layers, and standard attention -- which lack the capacity for fine-grained anatomical reasoning and multimodal alignment. Efforts to enhance MIR via external knowledge graphs \cite{li2023auxiliary} exist, but core architectural limitations persist. Our work builds on these limitations by introducing fresh and modern architectural units -- pre-normalization, latent attention, and gated feedforward specialized expert networks -- to address insufficient feature disambiguation and hierarchical alignment.

\subsection{Multiscale Image Feature Extraction}
Feature pyramid networks (FPNs) \cite{lin2017feature} are widely used in segmentation to model global and local patterns through stacked multi-resolution feature maps. MSNet \cite{zhang2023multi} and DeepPyramid$+$ \cite{ghamsarian2024deeppyramid} leverage deformable pyramids to enhance segmentation task. While these works are not directly focused on MIR, their principle of extracting anatomical features at varying resolutions is highly relevant. For instance, when conditioned on textual reports, these hierarchical visual maps can enrich anatomical grounding, which is crucial for structured report generation. Prior works like Bouslimi et al. \cite{bouslimi2025ai} only scratch the surface of FPN use in MIR, leaving multiscale encoding largely underexplored in this domain.

\subsection{Multimodal Alignment Mechanisms}
Conventional alignment in MIR relies on static self-attention \cite{vaswani2017attention}, which struggles with semantic alignment between visual and textual tokens. Some works explore reinforcement \cite{qinsong2022reinforced}, memory \cite{tao2024memory}, localized fusion mechanisms \cite{kim2021vilt, wang2023image}, and gated fusion mechanisms \cite{wang2023self}, but these often suffer from early entanglement or late-stage information loss. Further, these methods lack dynamic interaction modeling between image-text modalities. Our dual-branch latent attention and gated fusion modules address these issues enabling token-wise alignment refinement with visual evidence.

\subsection{Mixture of Expert Models}
Mixture-of-experts or MoE enables dynamic routing of tokens to specialized subnetworks \cite{shazeer2017outrageously, fedus2022switch}, enhancing both scalability and interoperability. In medical vision-language tasks, Med-MoE \cite{jiang2024med} applies MoE to visual QA, and M4oE \cite{jiang2024m4oe} introduces modality-specific routing for segmentation. However, MIR remains mostly untouched by MoE, particularly at the fine-grained level of report generation. Dense feedforward layers in prior MIR models compress all token interactions uniformly, while MoE allows token-dependent expert selection, improving specialization and interpretability. Our modulated MoE decoder addresses this gap by enabling dynamic expert activation, tailored to the semantic and anatomical complexity of the input.

\section{Method}
\subsection{Problem Formulation}
Let $I = \{i_1, ..., i_n\}$ denote input medical images from modality domain $\mathcal{I}$ (e.g., X-ray, CT, Fundus), and $D = \{d_1, ..., d_n\}$ be the corresponding free-text reports from output space $\mathcal{D}$ (unstructured clinical descriptions). The task is to learn a mapping $\mathcal{F}_\theta: \mathcal{I} \rightarrow \mathcal{D}$, parametrized by learnable weights $\theta$, such that $\hat{d}_n = \mathcal{F}_\theta(i_n)$ representing the predicted report for image $i_n$.

\subsection{Overview of the Approach}
The overall architecture of our proposed model, MicarVLMoE, is shown in \autoref{fig:arch}. It includes three main components designed to improve MIR through better feature extraction, alignment, and report generation. First, we use a multiscale vision encoder based on FPNs to capture fine-grained anatomical features at different resolutions, addressing the limitations of single-scale models. Second, a multihead dual-branch latent attention module learns latent bottleneck representations that align visual and textual information. To balance local anatomical structure encoding and global contextual understanding, our dual-branch latent attention integrates both RoPE-enhanced and non-RoPE-based heads. Third, a latent mixture-of-experts decoder is introduced to leverage a router-based MoE mechanism that activates specialized expert networks dynamically. We also include a gated fusion module that refines textual features evidenced on visual features via a gated combination network. These modules improve the coherence, structural organization, and clinical relevance of the generated reports, thereby enhancing fusion, alignment, and interpretability in MIR.

\subsection{Multiscale Vision Encoder}
\begin{figure}[!pbth]
   \centering
   \includegraphics[width=0.45\textwidth, height=1.5in]{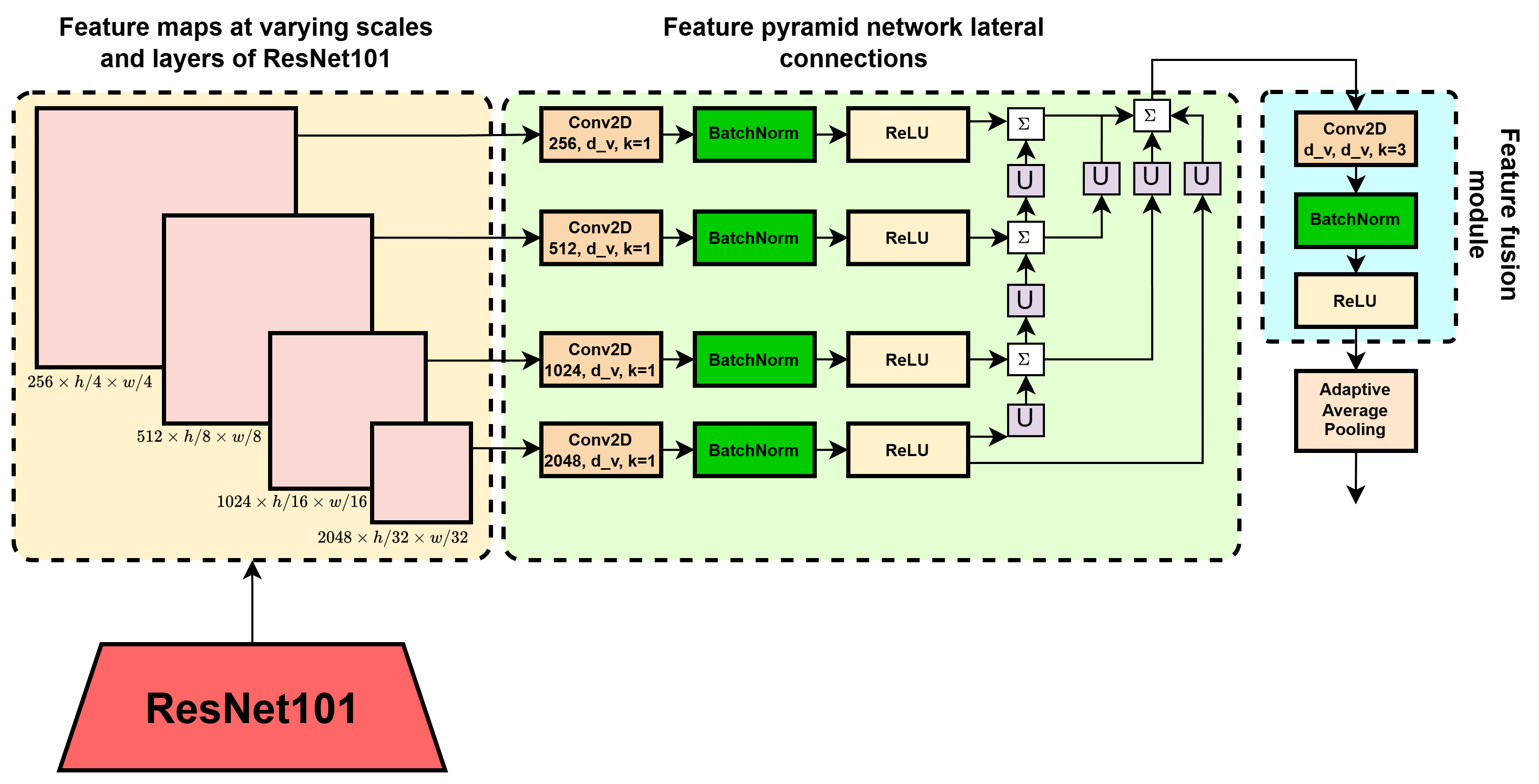}
   \caption{Illustration of multiscale vision encoder mechanism.}
   \label{fig:msve}
\end{figure}

To extract both low-level and high-level semantic features from medical images, we design a multiscale vision encoder (MSVE) module inspired by FPNs shown in \autoref{fig:msve}. This module processes medical images by capturing multi-scale anatomical structures, enhancing feature representations for both normal and abnormal regions. The core mechanism is described in detail as follows:

\noindent
\textbf{Feature Extraction.}~Given an input medical image $\mathbf{I} \in \mathbb{R}^{C \times H \times W}$, we use a CNN backbone $f = \{f_1, f_2, f_3, f_4\}$ to extract hierarchical feature maps at multiple scales: $\mathbf{C}_2 = f_1(\mathbf{I}) \in \mathbb{R}^{256 \times \frac{H}{4} \times \frac{W}{4}}$, $\mathbf{C}_3 = f_2(\mathbf{C}_2) \in \mathbb{R}^{512 \times \frac{H}{8} \times \frac{W}{8}}$, $\mathbf{C}_4 = f_3(\mathbf{C}_3) \in \mathbb{R}^{1024 \times \frac{H}{16} \times \frac{W}{16}}$, and $\mathbf{C}_5 = f_4(\mathbf{C}_4) \in \mathbb{R}^{2048 \times \frac{H}{32} \times \frac{W}{32}}$. Lower-level features retain fine-grained spatial detail, while deeper layers encode high-level semantic information.

\noindent
\textbf{Feature Pyramid Processing.}~To unify feature representations across scales, each feature map $\mathbf{C}_i$ is transformed using a $1 \times 1$ convolution followed by batch normalization and ReLU, denoted as $\phi(\cdot)$. The resulting pyramid features are computed as: $\mathbf{P}_5 = \phi(\mathbf{C}_5)$, $\mathbf{P}_4 = \phi(\mathbf{C}_4) + \lambda(\mathbf{P}_5, \mathbf{C}_4)$, $\mathbf{P}_3 = \phi(\mathbf{C}_3) + \lambda(\mathbf{P}_4, \mathbf{C}_3)$, and $\mathbf{P}_2 = \phi(\mathbf{C}_2) + \lambda(\mathbf{P}_3, \mathbf{C}_2)$, where $\lambda(\cdot,\cdot)$ denotes an interpolation function fusion ensuring spatial alignment.

\begin{figure}[!pbth]
    \centering
    \includegraphics[width=0.8\linewidth, height=1.15in]{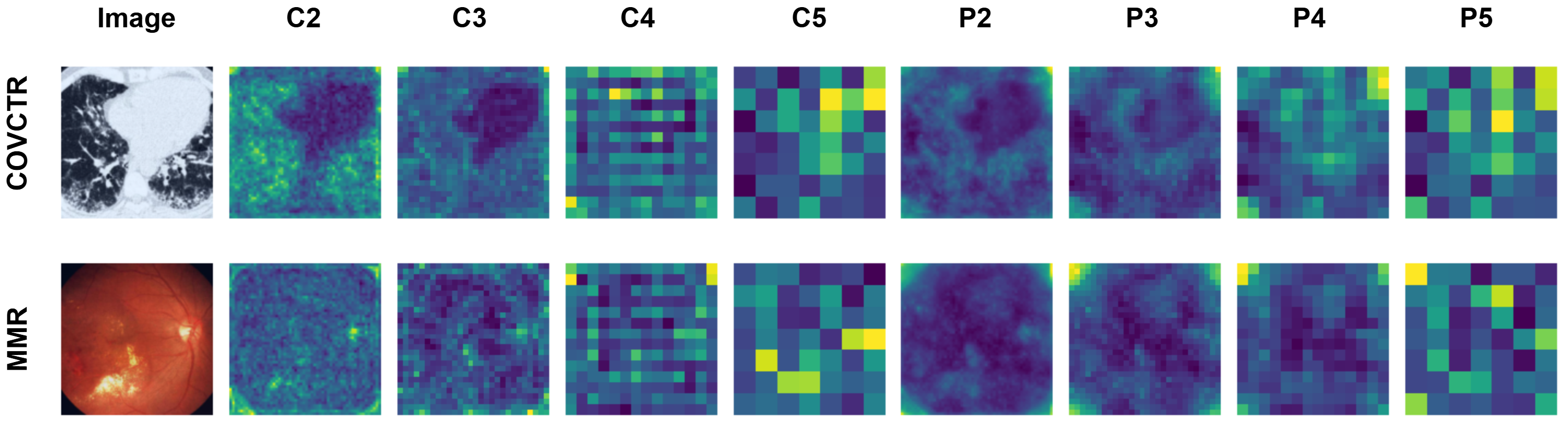}
    \caption{Visualization of hierarchical feature maps of the MSVE module.}
    \label{fig:msve-vis}
\end{figure}

\textbf{Multiscale Feature Fusion.}~The final feature representation $\mathbf{F} \in \mathbb{R}^{d_v \times \frac{H}{4} \times \frac{W}{4}}$ is obtained by aggregating all pyramid levels into a unified resolution: $\mathbf{F} = \Psi(\mathbf{P}_2 + \lambda(\mathbf{P}_3, \mathbf{P}_2) + \lambda(\mathbf{P}_4, \mathbf{P}_2) + \lambda(\mathbf{P}_5, \mathbf{P}_2))$, where $\Psi(\cdot)$ denotes a $3 \times 3$ convolution followed by batch normalization and ReLU to refine the fused multi-scale features. Finally, to generate a compact representation for downstream processing, we apply adaptive average pooling followed by reshaping to get $s_i$ image patch features $\mathbf{Z} \in \mathbb{R}^{s_i \times d_v}$. Then we project its dimension to $d_{\text{model}}$ via RMSNorm \cite{zhang2019root} and linear layer with output $\mathbf{Z} \in \mathbb{R}^{s_i \times d_{\text{model}}}$, which serves as input to the latent encoder in the MIR pipeline. MSVE's feature maps of varying scales are shown in \autoref{fig:msve-vis}.

\subsection{Latent Encoder}
\noindent
\textbf{Encoder Block.}~The encoder block refines the input feature representations using multihead dual-branch latent attention (MDLA) mechanism followed by gated feedforward network (FFN) \cite{touvron2023llama}. Given an input sequence $\mathbf{X} \in \mathbb{R}^{s \times d_{\text{model}}}$, the encoder block applies the following transformations: 
\begin{equation}
\begin{split}
    \mathbf{H}_{\text{att}} &= \text{MDLA}(\text{RMSNorm}(\mathbf{X})) + \mathbf{X} \\
    \mathbf{H}_{\text{ff}} &= \text{FFN}(\text{RMSNorm}(\mathbf{H}_{\text{att}})) + \mathbf{H}_{\text{att}}
\end{split}
\end{equation}
where $\mathbf{H}_{\text{ff}} \in \mathbb{R}^{s \times d_{\text{model}}}$.
This structure enables hierarchical feature refinement, ensuring that both local and global dependencies are captured effectively via the MDLA block before being passed to the next processing stage.

\begin{figure}[!pbth]
    \centering
    \includegraphics[width=0.40\textwidth, height=1.5in]{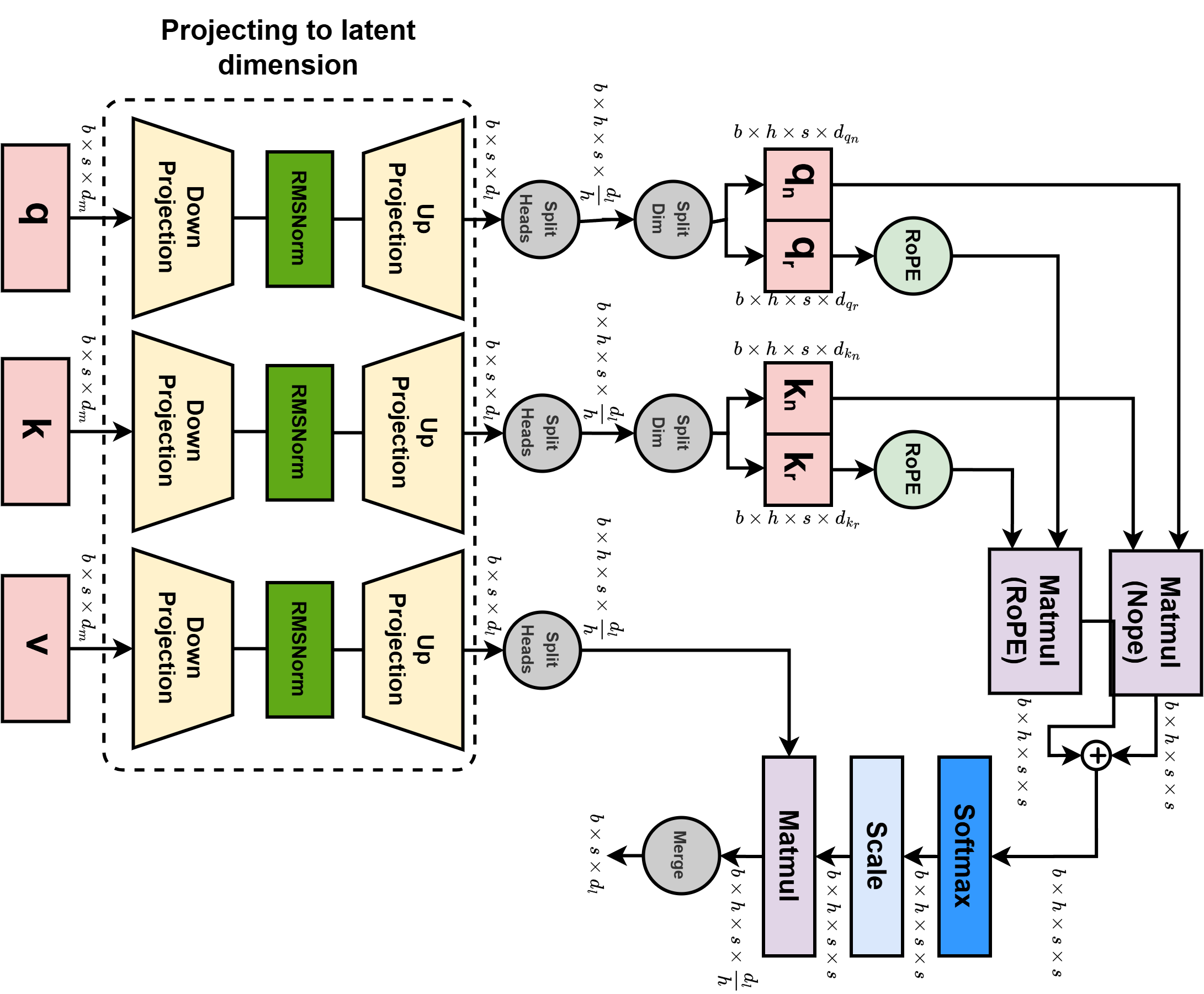}
    \caption{Visualization of multihead dual-branch latent attention mechanism. In the figure, the matrix size notations are as follows: $b$: batch size; $s$: sequence length; $d_{m}$: $d_{\text{model}}$; $d_l$: $d_{\text{latent}}$; $h$: number of heads; $d_{q_n}$,$d_{q_r}$,$d_{k_n}$,$d_{k_r}$: latent dimension of normal and rope query and key.}
    \label{fig:mdla}
\end{figure}

\noindent
\textbf{Multihead Dual-Branch Latent Attention.}~To refine medical image features for inter-contextual and spatial enhancement, we employ a multihead dual-branch latent attention (MDLA) mechanism illustrated in \autoref{fig:mdla}, inspired by DeepSeek-V3 \cite{liu2024deepseek}. Unlike conventional self-attention that operates directly on input tokens, MDLA introduces a learned latent bottleneck space, enabling feature interactions to be mediated through a compact representation. It encourages disentangled attention flow, enabling robust representation learning across spatial and contextual levels.

Given an input sequence of visual features $\mathbf{X} \in \mathbb{R}^{s \times d_{\text{model}}}$, query, key, and value projections are computed across $h$ heads as:
\begin{equation}
\begin{aligned}
    \mathbf{Q} &=& \left[\mathbf{q}_1;\mathbf{q}_2;...;\mathbf{q_h}\right] &=& \text{RMSNorm}(\mathbf{X} \mathbf{W}^{q}_{\text{down}}) \mathbf{W}^{q}_{\text{up}} \\
    \mathbf{K} &=& \left[\mathbf{k}_1;\mathbf{k}_2;...;\mathbf{k_h}\right] &=& \text{RMSNorm}(\mathbf{X} \mathbf{W}^{k}_{\text{down}}) \mathbf{W}^{k}_{\text{up}} \\
    \mathbf{V} &=& \left[\mathbf{v}_1;\mathbf{v}_2;...;\mathbf{v_h}\right] &=& \text{RMSNorm}(\mathbf{X} \mathbf{W}^{v}_{\text{down}}) \mathbf{W}^{v}_{\text{up}}
\end{aligned}
\end{equation}
where, $\mathbf{W}^{q,k,v}_{\text{down}} \in \mathbb{R}^{d_{\text{model}} \times d_{\text{model}} / h}$ and $\mathbf{W}^{q,k,v}_{\text{up}} \in \mathbb{R}^{d_{\text{model}} / h \times d_{\text{latent}}}$ are learned projections. The latent space allows intermediate attention computation and abstraction of anatomical features.

To incorporate positional cues while preserving abstraction, we decouple $\mathbf{Q}$ and $\mathbf{K}$ into two branches: (i) a RoPE-enhanced branch $(\mathbf{q}_r, \mathbf{k}_r)$ for capturing relative spatial dependencies and (ii) a standard branch $(\mathbf{q}_n, \mathbf{k}_n)$ without positional encoding (NoPE) for global feature reasoning. This dual-branch design jointly models local anatomical precision and global contextual alignment -- an approach not previously explored in medical vision-language tasks. The attention computation and output projection are defined as:
\begin{equation}
\begin{split}
    \mathbf{A}_h &= \frac{(\mathbf{q}_{n,h} \mathbf{k}_{n,h}^\top) + (\mathbf{q}_{r,h} \mathbf{k}_{r,h}^\top)}{\sqrt{d_{\text{latent}}}} \\ 
    \mathbf{Z}_h &= \text{softmax}(\mathbf{A}_h) \mathbf{v}_h \\
    \mathbf{Z} &= \left[\mathbf{Z}_1;\mathbf{Z}_2;...;\mathbf{Z}_h\right] \mathbf{W}_{\text{out}}
\end{split}
\end{equation}
where $\mathbf{A}_h \in \mathbb{R}^{s \times s}$, $\mathbf{Z} \in \mathbb{R}^{s \times d_{\text{model}}}$, and $\mathbf{W}_{\text{out}} \in \mathbb{R}^{d_{\text{latent}} \times d_{\text{model}}}$. This latent attention design improves spatial reasoning, positional grounding, and multimodal alignment -- particularly when used in cross-attention with textual features.

\subsection{Report Embedding}
The report embeddings are obtained by combining token embeddings with sinusoidal positional encodings $P$ for a given input report token sequence $\mathbf{X} \in \mathbb{R}^{s_t}$ as $\mathbf{E} = \text{Embedding}(\mathbf{X}) \cdot \sqrt{d_{\text{model}}} + \mathbf{P}$, where $\mathbf{E} \in \mathbb{R}^{s_t \times d_{\text{model}}}$.

\subsection{Latent MoE Decoder}
\noindent
\textbf{Decoder Block.}~The decoder block refines medical description representations by sequentially applying MDLA self-attention, gated fusion, and a mixture-of-experts mechanism. The MDLA self-attention mechanism enhances contextual dependencies between report tokens, ensuring both local coherence and global information aggregation. The MDLA cross-attention in gated fusion enables lexically diverse, clinically coherent reports that align textual reasoning with visual evidence. Given an input sequence $\mathbf{X} \in \mathbb{R}^{s_t \times d_{\text{model}}}$ and image features $\mathbf{I} \in \mathbb{R}^{s_i \times d_{\text{model}}}$, the decoder block performs the following transformations:
\begin{equation}
\begin{split}
    \mathbf{H}_{\text{sa}} &= \text{MDLA}(\text{RMSNorm}(\mathbf{X})) + \mathbf{X} \\
    \mathbf{H}_{\text{gf}} &= \text{GatedFusion}(\mathbf{H}_{\text{sa}}, \mathbf{I}) + \mathbf{H}_{\text{sa}} \\
    \mathbf{H}_{\text{moe}}, \mathcal{L}_{\text{b}} &= \text{MoE}(\text{RMSNorm}(\mathbf{H}_{\text{gf}})) + \mathbf{H}_{\text{gf}}
\end{split}
\end{equation}
where $\mathbf{H}_{\text{moe}} \in \mathbb{R}^{s_t \times d_{\text{model}}}$ and $\mathcal{L}_b$ is load balancing scalar loss explained in the subsequent sections.

\noindent
\textbf{Gated Fusion.}~The gated fusion module integrates image features with textual descriptions using MDLA-based cross-attention, a context-aware gating mechanism, and a fusion layer. Given text features $\mathbf{X} \in \mathbb{R}^{s_t \times d_{\text{model}}}$ and image features $\mathbf{I} \in \mathbb{R}^{s_i \times d_{\text{model}}}$, both modalities are first normalized and passed through MDLA cross-attention to produce image-aware textual features.
To inject global visual context, we extract summary statistics from $\mathbf{I}_{\mathrm{norm}}$ using both Mean and Max pooling, which jointly capture holistic (mean) and salient (max) image patterns -- providing complementary information for modulation. These are projected into a global context vector $\mathbf{G}$ to guide fusion. A gating function then blends attended image features and normalized text, selectively controlling visual influence at the token level. The final output is a fused representation informed by visual interactions as follows:
\begin{equation}
\begin{split}
    \mathbf{I}_{\mathrm{norm}}, \mathbf{X}_{\mathrm{norm}} &= \mathrm{RN}(\mathbf{I}), \,\mathrm{RN}(\mathbf{X}) \\
    \mathbf{I}_{\mathrm{att}} &= \mathrm{MDLA}(\mathbf{X}_{\mathrm{norm}}, \mathbf{I}_{\mathrm{norm}}) + \mathbf{X} \\
    \mathbf{G} &= \mathbf{W}_g \Bigl([\mathrm{Mean}(\mathbf{I}_{\mathrm{norm}});\mathrm{Max}(\mathbf{I}_{\mathrm{norm}})]\Bigr) + \mathbf{b}_g \\
    \mathbf{G}_{\mathrm{gate}} &= \sigma \Bigl(\mathrm{MLP}\bigl([\mathrm{RN}(\mathbf{I}_{\mathrm{att}}); \mathbf{X}_{\mathrm{norm}}]\bigr)\Bigr) \\
    \mathbf{F}_{\mathrm{gated}} &= \mathbf{G}_{\mathrm{gate}} \odot \mathrm{RN}(\mathbf{I}_{\mathrm{att}}) \;+\; (1 - \mathbf{G}_{\mathrm{gate}}) \odot \mathbf{X}_{\mathrm{norm}} \\
    \mathbf{Z} &= \mathrm{Fusion}\bigl([\mathbf{F}_{\mathrm{gated}};\mathbf{I}_{\mathrm{att}};\mathbf{G}]\bigr) \;+\; \mathbf{I}_{\mathrm{att}}
\end{split}
\end{equation}
where, $\sigma(\cdot)$ is the sigmoid activation and RN denotes RMSNorm. The output $\mathbf{Z} \in \mathbb{R}^{s_t \times d_{\text{model}}}$ is a refined report representation that captures localized image-text alignment, global anatomical context, and improved semantic grounding for MIR.

\noindent
\textbf{Mixture of Experts.}~The MoE module dynamically routes text tokens to specialized expert networks (gated FFNs), each responsible for modeling different semantic patterns. Given token features $\mathbf{X} \in \mathbb{R}^{s_t \times d_{\text{model}}}$, a gating mechanism computes token-wise expert scores via softmax, selects the top-$k$ experts per token, and combines their outputs. Expert specialization is not predefined (e.g., by anatomy or abnormalities), but instead emerges during training as experts compete to encode distinct clinical semantics. This allows the model to learn interpretable sub-functions aligned with medical reasoning. The MoE operations are defined as:
\begin{equation}
    \begin{aligned}
        \mathbf{S} &= \mathrm{softmax}(\mathbf{X} \mathbf{W}_g), \quad 
        \mathbf{S}_{\mathrm{top}}, \mathbf{I}_{\mathrm{top}} = \mathrm{TopK}(\mathbf{S}, k) \\
        \mathbf{Z} &= \sum_{i=1}^{k} \mathbf{S}_{\mathrm{top},i} \odot f_{\mathbf{I}_{\mathrm{top},i}}(\mathbf{X}), \quad
        \mathcal{L}_{\mathrm{b}} = \sum_{e=1}^{E} \bar{\mathbf{S}}_e \log \bar{\mathbf{S}}_e
    \end{aligned}
\end{equation}
where, $\mathbf{S} \in \mathbb{R}^{s_t \times E}$ denotes expert scores, $\mathbf{W}_g \in \mathbb{R}^{d_{\text{model}} \times E}$ is the gating projection, $\mathbf{Z}$ is the fused token representation, and $\mathcal{L}_{\mathrm{b}}$ is a load-balancing loss that promotes uniform expert utilization.

\subsection{Loss Function}
The decoder output is linearly projected to the vocabulary space and passed through a softmax to compute token probabilities. The training objective combines the cross-entropy loss for language modeling and a regularized MoE load-balancing loss accumulated over $N$ decoder layers as:
\begin{equation}
    \begin{aligned}
        \mathcal{L}_{\text{lang}} &= - \sum_{i=1}^{N} \log\bigl(\mathrm{pr}_{\theta}(s_{t_i} \mid s_{t_{1:i-1}}, I)\bigr), \\
        \mathcal{L}_{\text{balance}} &= \frac{1}{N} \sum_{n=1}^{N} \mathcal{L}_{\text{b}}^{(n)}, \quad 
        \mathcal{L}_{\text{total}} = \mathcal{L}_{\text{lang}} + \alpha\,\mathcal{L}_{\text{balance}}
    \end{aligned}
\end{equation}
where $\alpha$ controls the contribution of the MoE load-balancing term.

\section{Experiments}
\subsection{Datasets}
Experiments of this study utilize MIR datasets spanning multiple scans (X-ray, CT, MRI, fundus, OCT, etc) and anatomical sites (chest, head, neck, abdomen, eye, etc). The descriptions as are follows:

\noindent
\textbf{COVCTR.}~The COVCTR dataset \cite{li2023auxiliary} comprises 728 lung CT scan images, with 349 images representing COVID-19 cases and 379 depicting non-COVID cases. Each report includes sections on findings, impressions, terminology, and COVID-specific information.

\noindent
\textbf{MMRetinal.}~The MM-Retinal-V1 or MMR \cite{wu2024mm} dataset is a multi-modal retinal imaging dataset that includes 2,169 color fundus photography (CFP) cases, 1,947 fundus fluorescein angiography (FFA) cases, and 233 optical coherence tomography (OCT) cases. Each case consists of an image paired with descriptive texts available in both English and Chinese. The dataset is designed to facilitate research in multi-modal retinal disease analysis and report generation.

\noindent
\textbf{PGROSS.}~The PGROSS dataset \cite{jing2018automatic} contains descriptions from the Gross sub-collection of the PEIR digital library, totaling 7,442 image-caption pairs with associated tags. It spans 21 unique sub-categories related to lesion areas. Each caption in the dataset is structured as a single sentence.

\noindent
\textbf{ROCO.}~The ROCO \cite{pelka2018roco} dataset is a large-scale multimodal collection consisting of over 81,000 radiology images from the PubMed Central Open Access subset, each paired with descriptive captions. It includes a variety of imaging modalities such as X-ray, CT, MRI, ultrasound, and PET.

\subsection{Evaluation Metrics}
We evaluate model performance using standard MIR metrics: Bleu-n (B-n) \cite{papineni2002bleu}, Meteor \cite{banerjee2005meteor}, and Rouge \cite{lin2004rouge}, which are widely adopted for natural language generation. Additionally, we report RaTEScore \cite{zhao2024ratescore}, an entity-aware metric that assesses clinical accuracy by comparing key medical entities for relevance and semantic similarity.

\subsection{Implementation Details}
\noindent
\textbf{Splits and Preprocessing.}~Datasets are split into train/val/test as follows: COVCTR and MMR (8:1:1) \cite{tan2024medical}, PGROSS (7:1:2), and ROCO (official split). Reports are cleaned by removing de-identified and alphanumeric tokens, and augmented with special tokens $\langle$sos$\rangle$, $\langle$eos$\rangle$, $\langle$pad$\rangle$, and $\langle$unk$\rangle$. Tokens occurring $\geq$3 times are retained, covering 99\% of the target vocabulary. Rows with missing values are dropped. Maximum report lengths are set to 60 (MMR, ROCO), 80 (COVCTR), and 30 (PGROSS).

\noindent
\textbf{Training and Inference.}~We use ResNet101 \cite{he2016deep} pretrained on ImageNet1K \cite{deng2009imagenet} as the MSVE visual encoder, resizing images to 224$\times$224 and setting $d_v{=}2048$. Encoder-decoder layers are set to 2 for COVCTR and 3 for MMR, PGROSS, and ROCO, with 8 attention heads. We use SiLU activation for COVCTR, PGROSS, ROCO, and GELU for MMR. Across datasets, we set: $d_{\text{model}}{=}512$, $d_{\text{latent}}{=}768$, $q_{nk_n}{=}\,q_{rk_r}{=}48$, $d_{ff}{=}2048$, attention dropout to 0.12, block dropout to 0.1, and MoE with $E{=}8$, top-$k{=}2$. The loss weight $\alpha$ is 0.01. AdamW \cite{loshchilov2018decoupled} is used with learning rate $1\mathrm{e}{-4}$ (except MSVE: $5\mathrm{e}{-5}$), weight decay $5\mathrm{e}{-5}$, and StepLR scheduler ($\gamma{=}0.1$). Batch sizes are 16 (COVCTR, PGROSS, ROCO) and 32 (MMR). Training spans 100 epochs (COVCTR, MMR, PGROSS) and 30 epochs (ROCO). Inference uses beam search (width=3). All experiments run on Ubuntu 22.04, Python 3.10, PyTorch 2.1, with a single A100-80G GPU.

\noindent
\textbf{Baselines.}~We compare MicarVLMoE with state-of-the-art models: ST \cite{vinyals2015show}, SAT \cite{xu2015show}, CoAtt \cite{jing2018automatic}, ASGK \cite{li2023auxiliary}, AdaAtt \cite{lu2017knowing}, R2Gen \cite{chen2020generating}, MDAK/MDAKF \cite{tan2024medical}, AMANet \cite{yang2021automatic}, and Transformer \cite{vaswani2017attention}. For COVCTR, results are taken from \cite{tan2024medical} using the same split. For other datasets, we use results from original papers where possible, else we re-implement the method using the authors' official code and our preprocessing to ensure consistency and fair comparison.

\begin{table}[t]
\caption{Comparison with existing methods on the test set. Best results are bold, second-best underlined. "*" denotes reproduced results on our split. Avg-Bleu~$\Delta$ shows our model's improvement over others.}
\label{tab:res}
\centering
\tiny
\setlength{\tabcolsep}{2.5pt}
\renewcommand{\arraystretch}{1.2}
\begin{tabular}{|c|l|c|c|c|c|c|c|c|c|}
    \hline
    \textbf{Dataset} & \textbf{Method} 
    & \textbf{Bleu-1} & \textbf{Bleu-2} & \textbf{Bleu-3} & \textbf{Bleu-4} 
    & \textbf{Avg-Bleu} & \textbf{Avg-Bleu $\Delta$} 
    & \textbf{Meteor} & \textbf{Rouge-L} \\
    \hline

    \multirow{9}{*}{\textbf{COVCTR}} 
    & SAT \cite{xu2015show} 
      & 0.697 & 0.621 & 0.568 & 0.515 & 0.600 
      & 6.33\%\,\textcolor{green}{\(\uparrow\)} 
      & -- & 0.723 \\
    & CoAtt \cite{jing2018automatic} 
      & 0.709 & 0.645 & \underline{0.603} & 0.552 & 0.627 
      & 1.75\%\,\textcolor{green}{\(\uparrow\)} 
      & -- & \textbf{0.748} \\
    & ASGK \cite{li2023auxiliary} 
      & 0.712 & \underline{0.659} & \textbf{0.611} & \textbf{0.570} & \textbf{0.638} 
      & 0.00\% 
      & -- & \underline{0.746} \\
    & AdaAtt \cite{lu2017knowing} 
      & 0.676 & 0.633 & 0.596 & 0.514 & 0.605
      & 5.45\%\,\textcolor{green}{\(\uparrow\)} 
      & -- & 0.726 \\
    & Transformer$^{*}$ \cite{vaswani2017attention} 
      & 0.713 & 0.631 & 0.568 & 0.515 & 0.607 
      & 5.11\%\,\textcolor{green}{\(\uparrow\)} 
      & \underline{0.655} & 0.656 \\
    & R2Gen$^{*}$ \cite{chen2020generating} 
      & \underline{0.736} & 0.657 & 0.593 & 0.539 & \underline{0.631} 
      & 1.11\%\,\textcolor{green}{\(\uparrow\)} 
      & 0.398 & 0.701 \\
    & MDAK \cite{tan2024medical} 
      & 0.723 & 0.652 & 0.586 & \underline{0.545} & 0.626 
      & 1.92\%\,\textcolor{green}{\(\uparrow\)} 
      & 0.403 & 0.676 \\
    & MDAKF \cite{tan2024medical} 
      & 0.726 & 0.651 & 0.583 & 0.539 & 0.625 
      & 2.08\%\,\textcolor{green}{\(\uparrow\)} 
      & 0.401 & 0.683 \\
    \cline{2-10}
    & \textbf{Ours} 
      & \textbf{0.744} & \textbf{0.662} & 0.599 & \underline{0.545} & \textbf{0.638} 
      & --- 
      & \textbf{0.690} & 0.692 \\
    \hline

    \multirow{7}{*}{\textbf{PGROSS}} 
    & AdaAtt \cite{lu2017knowing} 
      & 0.210 & 0.149 & 0.111 & 0.089 & 0.140
      & 55.71\%\,\textcolor{green}{\(\uparrow\)}
      & 0.241 & 0.102 \\
    & AMANet \cite{yang2021automatic} 
      & 0.200 & 0.144 & 0.112 & 0.093 & 0.137 
      & 59.12\%\,\textcolor{green}{\(\uparrow\)}
      & 0.247 & 0.100 \\
    & ST$^{*}$ \cite{vinyals2015show} 
      & 0.097 & 0.064 & 0.044 & 0.020 & 0.056 
      & 289.29\%\,\textcolor{green}{\(\uparrow\)} 
      & 0.145 & 0.128 \\
    & SAT$^{*}$ \cite{xu2015show} 
      & 0.208 & 0.130 & 0.089 & 0.055 & 0.121 
      & 80.17\%\,\textcolor{green}{\(\uparrow\)}
      & 0.212 & 0.214 \\
    & Transformer$^{*}$ \cite{vaswani2017attention} 
      & 0.257 & 0.179 & 0.132 & 0.093 & 0.165
      & 32.12\%\,\textcolor{green}{\(\uparrow\)}
      & \underline{0.282} & 0.268 \\
    & R2Gen$^{*}$ \cite{chen2020generating} 
      & \textbf{0.340} & \underline{0.229} & \underline{0.171} & \underline{0.121} & \underline{0.215} 
      & 1.40\%\,\textcolor{green}{\(\uparrow\)}
      & 0.155 & \textbf{0.385} \\ 
    \cline{2-10}
    & \textbf{Ours} 
      & \underline{0.321} & \textbf{0.236} & \textbf{0.182} & \textbf{0.132} & \textbf{0.218} 
      & ---
      & \textbf{0.287} & \underline{0.302}  \\
    \hline

    \multirow{5}{*}{\textbf{MMR}} 
    & ST$^{*}$ \cite{vinyals2015show} 
      & 0.110 & 0.069 & 0.043 & 0.026 & 0.062
      & 241.94\%\,\textcolor{green}{\(\uparrow\)}
      & 0.138 & 0.132 \\
    & SAT$^{*}$ \cite{xu2015show} 
      & 0.215 & 0.134 & 0.090 & 0.058 & 0.124 
      & 70.97\%\,\textcolor{green}{\(\uparrow\)}
      & 0.227 & 0.202 \\
    & Transformer$^{*}$ \cite{vaswani2017attention} 
      & 0.242 & 0.163 & 0.115 & 0.083 & 0.151 
      & 40.40\%\,\textcolor{green}{\(\uparrow\)}
      & \underline{0.282} & 0.234 \\
    & R2Gen$^{*}$ \cite{chen2020generating} 
      & \underline{0.307} & \underline{0.210} & \underline{0.150} & \underline{0.110} & \underline{0.194} 
      & 9.28\%\,\textcolor{green}{\(\uparrow\)}
      & 0.128 & \textbf{0.310} \\ 
    \cline{2-10}
    & \textbf{Ours} 
      & \textbf{0.321} & \textbf{0.225} & \textbf{0.168} & \textbf{0.132} & \textbf{0.212}
      & ---
      & \textbf{0.333} & \underline{0.283} \\
    \hline 

    \multirow{4}{*}{\textbf{ROCO}} 
    & ST$^{*}$ \cite{vinyals2015show} 
      & 0.067 & 0.033 & 0.014 & 0.006 & 0.030
      & 113.33\%\,\textcolor{green}{\(\uparrow\)}
      & 0.108 & 0.087 \\
    & SAT$^{*}$ \cite{xu2015show} 
      & 0.109 & 0.052 & 0.022 & 0.009 & 0.048
      & 33.33\%\,\textcolor{green}{\(\uparrow\)}
      & 0.117 & 0.111 \\
    & Transformer$^{*}$ \cite{vaswani2017attention} 
      & \underline{0.110} & \underline{0.055} & \underline{0.028} & \underline{0.014} & \underline{0.052} 
      & 23.08\%\,\textcolor{green}{\(\uparrow\)}
      & \underline{0.148} & \underline{0.123} \\
    \cline{2-10}
    & \textbf{Ours} 
      & \textbf{0.136} & \textbf{0.069} & \textbf{0.034} & \textbf{0.018} & \textbf{0.064}
      & ---
      & \textbf{0.177} & \textbf{0.144} \\ 
    \hline
\end{tabular}
\end{table}

\begin{figure*}[!pbth]
    \centering
    \includegraphics[width=\linewidth,height=3in]{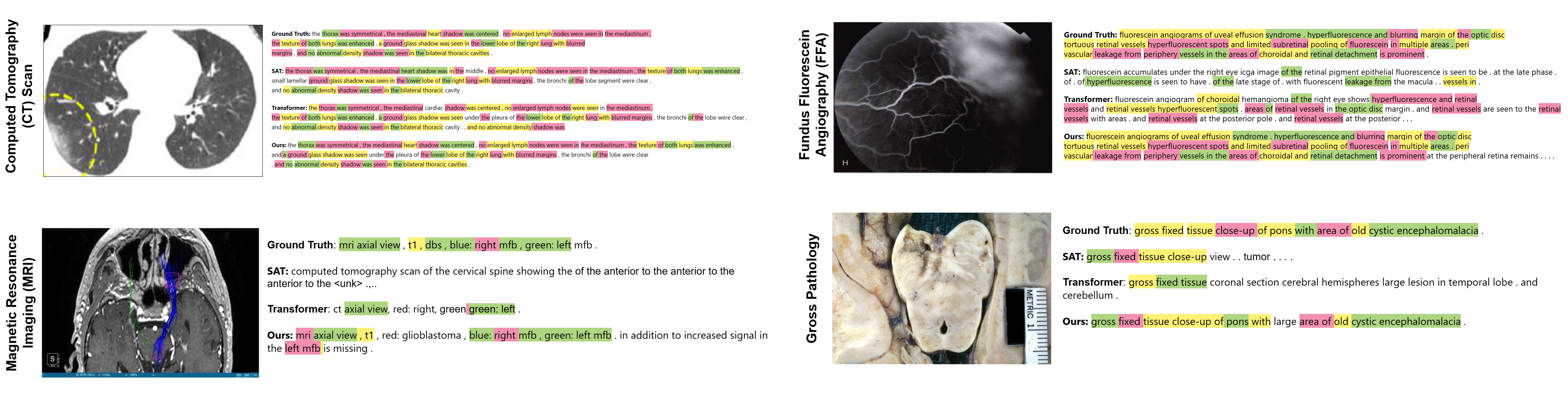}
    \caption{Example of generated reports compared between SAT, Transformer, and our (MicarVLMoE) model. Highlighted colors depict medically relevant findings matched with the ground truth reports/captions.}
    \label{fig:qual}
\end{figure*}

\section{Results}\label{sec:Results}

\subsection{SOTA Quantitative Analysis}
As shown in \autoref{tab:res}, MicarVLMoE outperforms or matches all prior methods across COVCTR, PGROSS, MMR, and ROCO, consistently ranking first or tying across key metrics. On COVCTR, it achieves the best Bleu-1/2 and Meteor scores. For PGROSS, while slightly behind R2Gen on Bleu-1, it leads on all other metrics. On MMR, it sets new state-of-the-art scores across all Bleu and Meteor variants. On ROCO, it surpasses all baselines in Bleu, Meteor, and Rouge-L, demonstrating strong generalization across diverse anatomical regions and imaging types.

As shown in \autoref{tab:ratescore}, MicarVLMoE also achieves the highest RaTEScore across all datasets. This gain stems from its MDLA and gated fusion, enabling finer visual-text alignment compared to vanilla attention (SAT) and standard multi-head attention (Transformer). These results validate RQ1 and RQ2, confirming the model’s effectiveness across imaging modalities and its ability to maintain accurate, clinically grounded report generation.

\begin{table}[!tb]
\centering
\caption{RaTEScore performance (mean\,$\pm$\,std) across attention-based models.}
\label{tab:ratescore}
\scriptsize
\renewcommand{\arraystretch}{1.1}
    \begin{tabular}{|c|c|c|c|}
    \hline
    \textbf{Dataset} & \textbf{SAT} & \textbf{Transformer} & \textbf{Ours} \\
    \hline
    COVCTR  & 0.729\,$\pm$\,0.107 & 0.740\,$\pm$\,0.118 & \textbf{0.745\,$\pm$\,0.116} \\
    \hline
    MMR     & 0.439\,$\pm$\,0.110 & 0.485\,$\pm$\,0.138 & \textbf{0.505\,$\pm$\,0.151} \\
    \hline
    PGROSS  & 0.317\,$\pm$\,0.148 & 0.395\,$\pm$\,0.163 & \textbf{0.397\,$\pm$\,0.177} \\
    \hline
    ROCO    & 0.269\,$\pm$\,0.135 & 0.348\,$\pm$\,0.134 & \textbf{0.358\,$\pm$\,0.131} \\
    \hline
    \end{tabular}
\end{table}

\subsection{Ablation Study}
As shown in \autoref{tab:ab-model}, each component of MicarVLMoE contributes to overall performance. Removing the load balancing loss (lb) reduces average scores from 0.638 to 0.609 (COVCTR) and 0.212 to 0.190 (MMR), confirming its role in stabilizing expert usage. Excluding the MoE module yields similar drops (0.628/0.189), highlighting the value of expert-based specialization. Removing the latent encoder (lenc) further lowers performance (0.608/0.180), showing its impact on learning deeper representations. The largest decline occurs when removing the multiscale vision encoder (msve), dropping to 0.595 (COVCTR) and 0.164 (MMR), underscoring the importance of multi-resolution feature extraction. These results support RQ1, validating the contribution of each module towards effective multimodal alignment.

\begin{table}[!pbth]
\caption{Ablation results on model components. "Lb", "moe", "lenc", and "msve" denote load balancing, mixture-of-experts, latent encoder, and multiscale vision encoder.}
\label{tab:ab-model}
\centering
\scriptsize
\setlength{\tabcolsep}{3pt}
\renewcommand{\arraystretch}{0.9}
\begin{tabular}{|c|l|c|c|c|c|c|}
\hline
\textbf{Dataset} & \textbf{Model} & \textbf{B-1} & \textbf{B-2} & \textbf{B-3} & \textbf{B-4} & \textbf{Avg-B} \\
\hline

\multirow{5}{*}{\textbf{COVCTR}} 
& \textbf{Full model} & \textbf{0.744} & \textbf{0.662} & \textbf{0.599} & \textbf{0.545} & \textbf{0.638} \\
& w/o lb loss         & 0.721 & 0.635 & 0.568 & 0.511 & 0.609 \\
& w/o moe             & 0.737 & 0.652 & 0.588 & 0.533 & 0.628 \\
& w/o lenc            & 0.717 & 0.632 & 0.568 & 0.513 & 0.608 \\
& w/o msve            & 0.709 & 0.620 & 0.553 & 0.497 & 0.595 \\
\hline

\multirow{5}{*}{\textbf{MMR}} 
& \textbf{Full model} & \textbf{0.321} & \textbf{0.225} & \textbf{0.168} & \textbf{0.132} & \textbf{0.212} \\
& w/o lb loss         & 0.295 & 0.202 & 0.148 & 0.113 & 0.190 \\
& w/o moe             & 0.297 & 0.202 & 0.146 & 0.111 & 0.189 \\
& w/o lenc            & 0.281 & 0.190 & 0.140 & 0.107 & 0.180 \\
& w/o msve            & 0.267 & 0.176 & 0.123 & 0.090 & 0.164 \\
\hline
\end{tabular}
\end{table}

\subsection{Hyperparameter Analysis}
\begin{figure}[!pbth]
    \centering
    \includegraphics[width=\linewidth, height=3.5in]{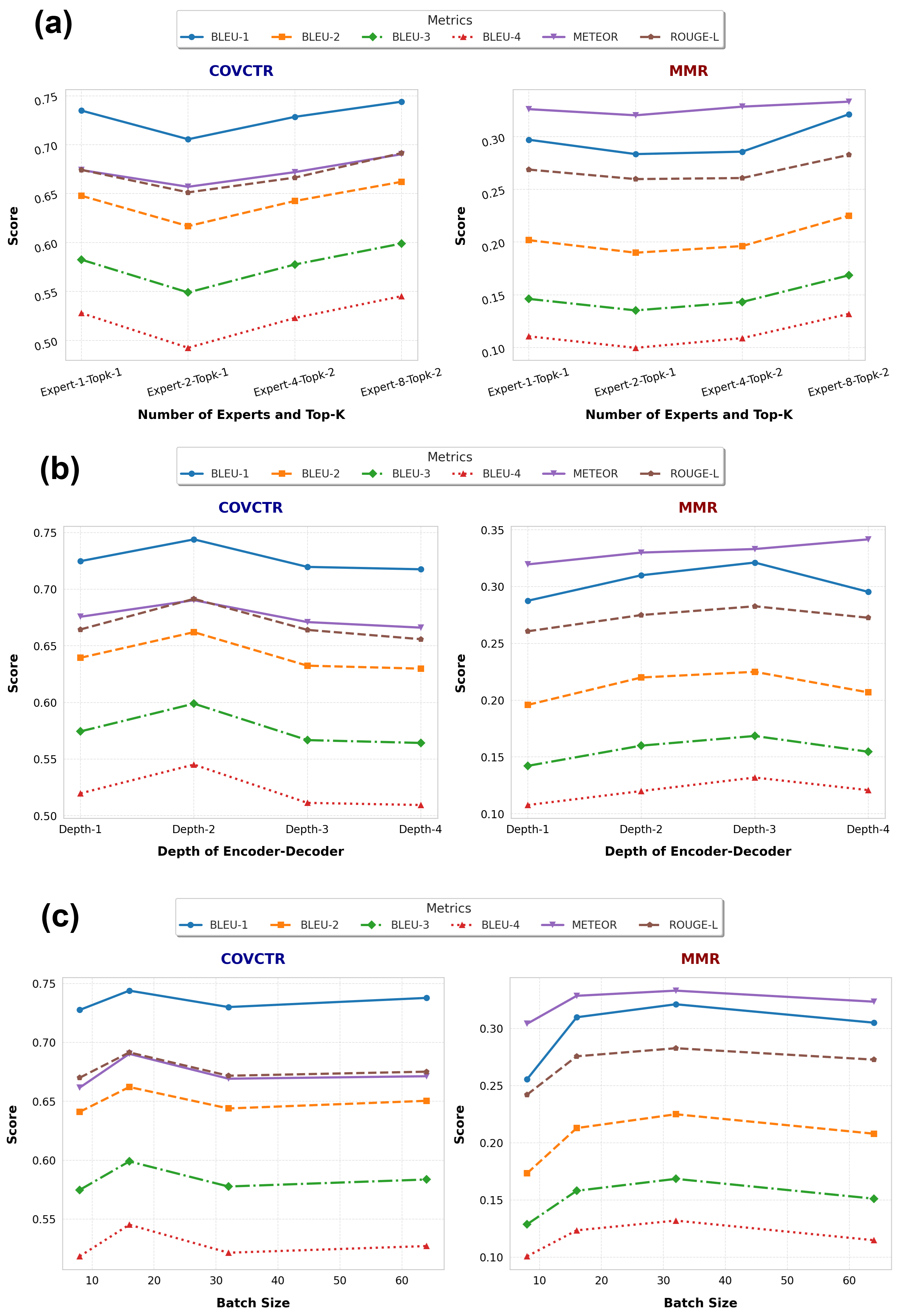}
    \caption{Effect of (a) Number of experts in the MoE layer with active topk experts; (b) Depth of encoder-decoder layers; (c) Batch size -- on COVCTR and MMR dataset.}
    \label{fig:hp}
\end{figure}

We analyze the effect of (1) number of experts, (2) encoder-decoder depth, and (3) batch size on COVCTR and MMR. As shown in \autoref{fig:hp}\textcolor{blue}{a}, performance improves with more experts, peaking at Expert-8-TopK-2, while lower diversity (Expert-2-TopK-1) reduces performance. In \autoref{fig:hp}\textcolor{blue}{b}, Depth-2 gives the best results, while deeper settings lead to degradation -- likely due to overfitting or optimization challenges. \autoref{fig:hp}\textcolor{blue}{c} shows batch sizes of 20-30 yield optimal performance; extremes negatively impact generalization. These results highlight the importance of tuning MoE size, model depth, and batch size for optimal MIR outcomes.

\subsection{Qualitative Analysis}
\noindent 
\textbf{Generated Report Analysis.}~\autoref{fig:qual} presents qualitative comparisons across four imaging modalities. MicarVLMoE generates clinically precise, structured reports that closely align with ground truth, outperforming SAT and Transformer baselines. For instance, in CT and MRI cases, our model captures detailed anatomical terms (e.g., ``ground glass shadow", ``mri axial view") and localizes findings like ``left mfb" more accurately. In fundus FFA, it correctly identifies multi-area vascular leakage, while SAT and Transformer miss key findings or introduce fragmented text. In gross pathology, our model effectively recovers the full pathology description, unlike SAT's incomplete output or the Transformer’s generic phrasing. Minor errors persist in rare phrasing or subtle anatomical localization, but overall, MicarVLMoE demonstrates superior clinical grounding, detail preservation, and cross-modal alignment.

\begin{figure}[!pbth]
    \centering
    \includegraphics[width=\linewidth]{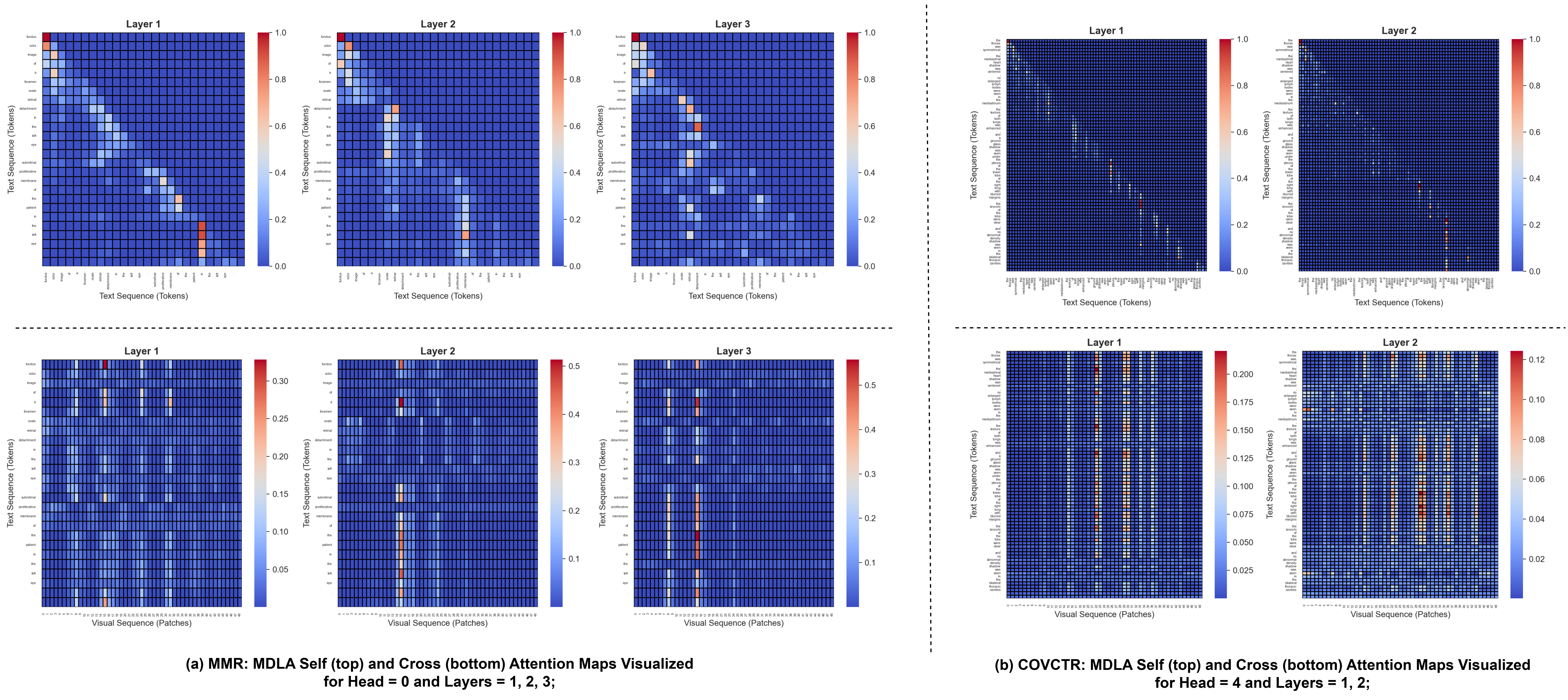}
    \caption{Visualization of MDLA attention maps - both self and cross attentions for (a) MMR; (b) COVCTR.}
    \label{fig:mdla-vis}
\end{figure}

\noindent
\textbf{MDLA Attention Maps Visualization Analysis.}~\autoref{fig:mdla-vis} illustrates MDLA's multi-modal attention behavior on MMR and COVCTR. Self-attention maps (top) show strong intra-token interactions in deeper layers, while cross-attention maps (bottom) highlight precise alignment between text and visual patches. Deeper layers exhibit more focused activations, reflecting refined semantic fusion. These patterns demonstrate MDLA’s role in enhancing multimodal alignment and contextual understanding in MIR.

\begin{figure}[!pbth]
    \centering
    \includegraphics[width=0.8\linewidth, height=1.8in]{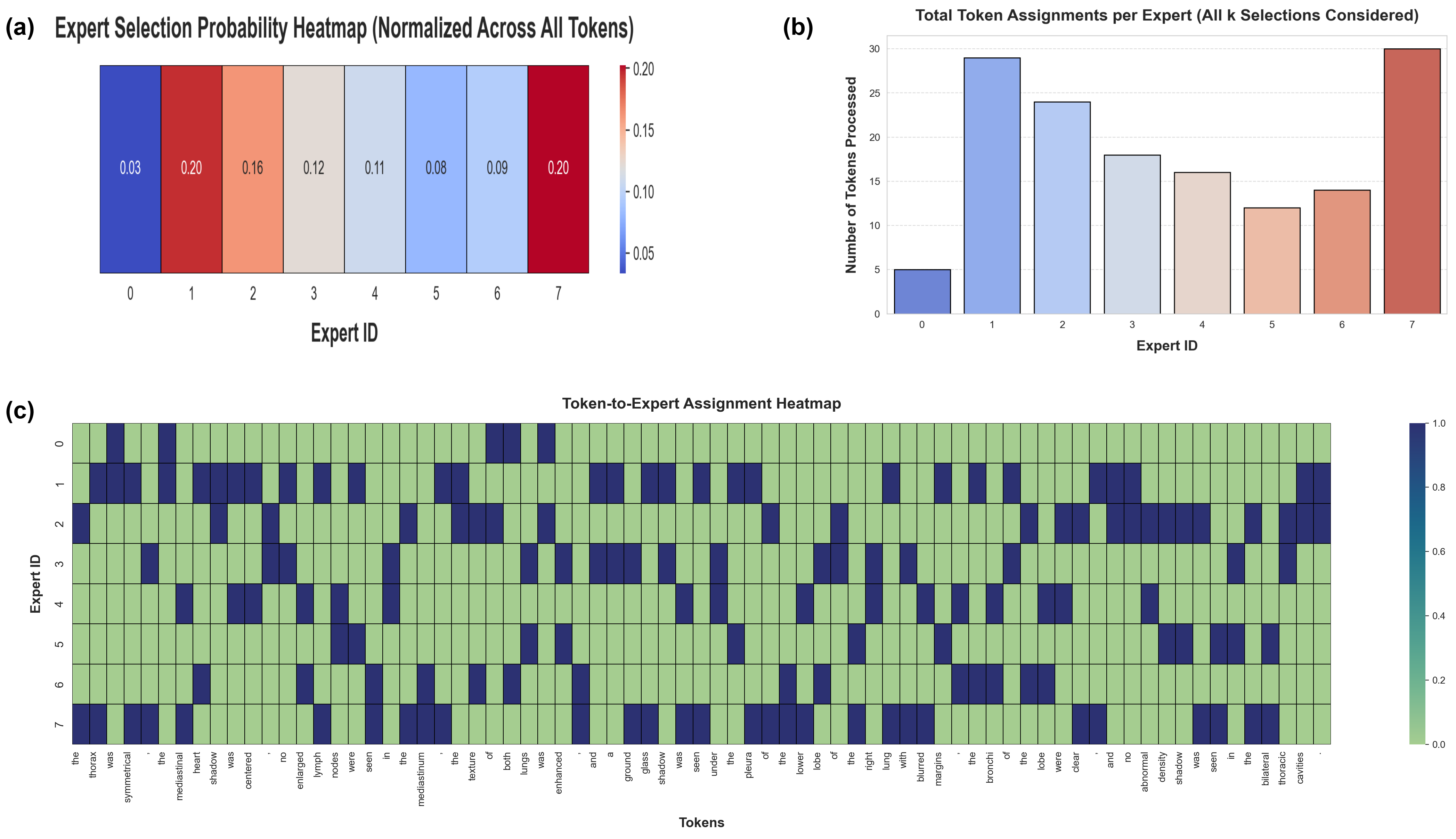}
    \caption{MoE experts visualized for an example report from COVCTR; (a) Expert selection probability heatmap; (b) Token assignment i.e. load balancing of tokens across experts; (c) Distribution of experts assigned to sequential tokens.}
    \label{fig:moe-vis}
\end{figure}

\noindent
\textbf{MoE Visualization Analysis.}~\autoref{fig:moe-vis} illustrates expert utilization on COVCTR. The expert selection heatmap (a) shows balanced activation with signs of specialization across experts. Token distribution (b) confirms effective load balancing without expert overload. The token-to-expert assignment map (c) reveals meaningful routing patterns: anatomical entities like ``thorax", ``mediastinal", and ``bronchi" are primarily processed by Experts 1 and 7; observational findings such as ``shadow", ``ground glass" engage Experts 2 and 3; while uncertainty or absence descriptors like ``no", ``clear" activate different experts, particularly Experts 4 and 6. This specialization across semantic categories enables more structured, clinically coherent, and explainable report generation. Overall, these patterns highlight the advantage of expert-driven modeling over dense FFNs in MIR tasks.

\section{Conclusion}
We presented MicarVLMoE, a gated cross-aligned vision-language mixture-of-experts model for medical image captioning and report generation. By combining multihead dual-branch latent attention, gated fusion, and MoE-based specialization, our approach improves multimodal alignment, feature representation, and interpretability. MicarVLMoE also broadens MIR research by establishing benchmarks across CT, retinal, MRI, and pathology datasets. Experiments and ablations confirm state-of-the-art clinical accuracy. In future work, we aim to integrate entity-based knowledge graphs into MDLA for richer contextual grounding, scale the model further ($\geq$1B), and explore joint training across diverse modalities with human evaluations for clinical validation.

\bibliographystyle{IEEEtran}
\bibliography{ref}

\begin{thebibliography}{10}
\providecommand{\url}[1]{#1}
\csname url@samestyle\endcsname
\providecommand{\newblock}{\relax}
\providecommand{\bibinfo}[2]{#2}
\providecommand{\BIBentrySTDinterwordspacing}{\spaceskip=0pt\relax}
\providecommand{\BIBentryALTinterwordstretchfactor}{4}
\providecommand{\BIBentryALTinterwordspacing}{\spaceskip=\fontdimen2\font plus
\BIBentryALTinterwordstretchfactor\fontdimen3\font minus \fontdimen4\font\relax}
\providecommand{\BIBforeignlanguage}[2]{{%
\expandafter\ifx\csname l@#1\endcsname\relax
\typeout{** WARNING: IEEEtran.bst: No hyphenation pattern has been}%
\typeout{** loaded for the language `#1'. Using the pattern for}%
\typeout{** the default language instead.}%
\else
\language=\csname l@#1\endcsname
\fi
#2}}
\providecommand{\BIBdecl}{\relax}
\BIBdecl

\bibitem{alfarghaly2021automated}
O.~Alfarghaly, R.~Khaled, A.~Elkorany, M.~Helal, and A.~Fahmy, ``Automated radiology report generation using conditioned transformers,'' \emph{Informatics in Medicine Unlocked}, vol.~24, p. 100557, 2021.

\bibitem{yu2024long}
L.~Yu, X.~Wang, B.~Deng, and C.~Ma, ``The long-term memory transformer with multimodal fusion for radiology report generation,'' in \emph{2024 International Joint Conference on Neural Networks (IJCNN)}, 2024, pp. 1--6.

\bibitem{izhar2025engaging}
A.~Izhar, N.~Idris, and N.~Japar, ``Engaging preference optimization alignment in large language model for continual radiology report generation: A hybrid approach,'' \emph{Cognitive Computation}, vol.~17, no.~1, p.~53, 2025.

\bibitem{liu2019clinically}
G.~Liu, T.-M.~H. Hsu, M.~McDermott, W.~Boag, W.-H. Weng, P.~Szolovits, and M.~Ghassemi, ``Clinically accurate chest x-ray report generation,'' in \emph{Machine Learning for Healthcare Conference}, 2019, pp. 249--269.

\bibitem{aksoy2023radiology}
N.~Aksoy, N.~Ravikumar, and A.~F. Frangi, ``Radiology report generation using transformers conditioned with non-imaging data,'' in \emph{Medical Imaging 2023: Imaging Informatics for Healthcare, Research, and Applications}, vol. 12469, 2023, pp. 146--153.

\bibitem{xu2023vision}
D.~Xu, H.~Zhu, Y.~Huang, Z.~Jin, W.~Ding, H.~Li, and M.~Ran, ``Vision-knowledge fusion model for multi-domain medical report generation,'' \emph{Information Fusion}, vol.~97, p. 101817, 2023.

\bibitem{vaswani2017attention}
A.~Vaswani, N.~Shazeer, N.~Parmar, J.~Uszkoreit, L.~Jones, A.~N. Gomez, {\L}.~Kaiser, and I.~Polosukhin, ``Attention is all you need,'' \emph{Advances in Neural Information Processing Systems}, vol.~30, 2017.

\bibitem{chen2020generating}
Z.~a. Chen, ``Generating radiology reports via memory-driven transformer,'' in \emph{Proceedings of the 2020 Conference on Empirical Methods in Natural Language Processing (EMNLP)}, B.~a. Webber, Ed.\hskip 1em plus 0.5em minus 0.4em\relax Association for Computational Linguistics, nov 2020, pp. 1439--1449.

\bibitem{kim2021vilt}
W.~Kim, B.~Son, and I.~Kim, ``Vilt: Vision-and-language transformer without convolution or region supervision,'' in \emph{International Conference on Machine Learning}, 2021, pp. 5583--5594.

\bibitem{wang2023image}
W.~Wang, H.~Bao, L.~Dong, J.~Bjorck, Z.~Peng, Q.~Liu, K.~Aggarwal, O.~K. Mohammed, S.~Singhal, S.~Som \emph{et~al.}, ``Image as a foreign language: Beit pretraining for vision and vision-language tasks,'' in \emph{Proceedings of the IEEE/CVF Conference on Computer Vision and Pattern Recognition}, 2023, pp. 19\,175--19\,186.

\bibitem{li2023auxiliary}
M.~Li, R.~Liu, F.~Wang, X.~Chang, and X.~Liang, ``Auxiliary signal-guided knowledge encoder-decoder for medical report generation,'' \emph{World Wide Web}, vol.~26, no.~1, pp. 253--270, 2023.

\bibitem{lin2017feature}
T.-Y. Lin, P.~Doll{\'a}r, R.~Girshick, K.~He, B.~Hariharan, and S.~Belongie, ``Feature pyramid networks for object detection,'' in \emph{Proceedings of the IEEE Conference on Computer Vision and Pattern Recognition}, 2017, pp. 2117--2125.

\bibitem{zhang2023multi}
B.~Zhang, Y.~Wang, C.~Ding, Z.~Deng, L.~Li, Z.~Qin, Z.~Ding, L.~Bian, and C.~Yang, ``Multi-scale feature pyramid fusion network for medical image segmentation,'' \emph{International Journal of Computer Assisted Radiology and Surgery}, vol.~18, no.~2, pp. 353--365, 2023.

\bibitem{ghamsarian2024deeppyramid}
N.~Ghamsarian, S.~Wolf, M.~Zinkernagel, K.~Schoeffmann, and R.~Sznitman, ``Deeppyramid+: Medical image segmentation using pyramid view fusion and deformable pyramid reception,'' \emph{International Journal of Computer Assisted Radiology and Surgery}, vol.~19, no.~5, pp. 851--859, 2024.

\bibitem{bouslimi2025ai}
R.~Bouslimi, H.~Trabelsi, W.~B.~A. Karaa, and H.~Hedhli, ``Ai-driven radiology report generation for traumatic brain injuries,'' \emph{Journal of Imaging Informatics in Medicine}, pp. 1--16, 2025.

\bibitem{qinsong2022reinforced}
H.~a. Qin, ``Reinforced cross-modal alignment for radiology report generation,'' in \emph{Findings of the Association for Computational Linguistics: ACL 2022}, S.~a. Muresan, Ed.\hskip 1em plus 0.5em minus 0.4em\relax Association for Computational Linguistics, may 2022, pp. 448--458.

\bibitem{tao2024memory}
Y.~Tao, L.~Ma, J.~Yu, and H.~Zhang, ``Memory-based cross-modal semantic alignment network for radiology report generation,'' \emph{IEEE Journal of Biomedical and Health Informatics}, 2024.

\bibitem{wang2023self}
Y.~Wang, K.~Wang, X.~Liu, T.~Gao, J.~Zhang, and G.~Wang, ``Self adaptive global-local feature enhancement for radiology report generation,'' in \emph{2023 IEEE International Conference on Image Processing (ICIP)}, 2023, pp. 2275--2279.

\bibitem{shazeer2017outrageously}
N.~Shazeer, A.~Mirhoseini, K.~Maziarz, A.~Davis, Q.~Le, G.~Hinton, and J.~Dean, ``Outrageously large neural networks: The sparsely-gated mixture-of-experts layer,'' \emph{arXiv preprint arXiv:1701.06538}, 2017.

\bibitem{fedus2022switch}
W.~Fedus, B.~Zoph, and N.~Shazeer, ``Switch transformers: Scaling to trillion parameter models with simple and efficient sparsity,'' \emph{Journal of Machine Learning Research}, vol.~23, no. 120, pp. 1--39, 2022.

\bibitem{jiang2024med}
S.~Jiang, T.~Zheng, Y.~Zhang, Y.~Jin, L.~Yuan, and Z.~Liu, ``Med-moe: Mixture of domain-specific experts for lightweight medical vision-language models,'' in \emph{Findings of the Association for Computational Linguistics: EMNLP 2024}, 2024, pp. 3843--3860.

\bibitem{jiang2024m4oe}
Y.~Jiang and Y.~Shen, ``M4oe: A foundation model for medical multimodal image segmentation with mixture of experts,'' in \emph{International Conference on Medical Image Computing and Computer-Assisted Intervention}, 2024, pp. 621--631.

\bibitem{zhang2019root}
B.~Zhang and R.~Sennrich, ``Root mean square layer normalization,'' \emph{Advances in Neural Information Processing Systems}, vol.~32, 2019.

\bibitem{touvron2023llama}
H.~Touvron, L.~Martin, K.~Stone, P.~Albert, A.~Almahairi, Y.~Babaei, N.~Bashlykov, S.~Batra, P.~Bhargava, S.~Bhosale \emph{et~al.}, ``Llama 2: Open foundation and fine-tuned chat models,'' \emph{arXiv preprint arXiv:2307.09288}, 2023.

\bibitem{liu2024deepseek}
A.~Liu, B.~Feng, B.~Xue, B.~Wang, B.~Wu, C.~Lu, C.~Zhao, C.~Deng, C.~Zhang, C.~Ruan \emph{et~al.}, ``Deepseek-v3 technical report,'' \emph{arXiv preprint arXiv:2412.19437}, 2024.

\bibitem{wu2024mm}
R.~Wu, C.~Zhang, J.~Zhang, Y.~Zhou, T.~Zhou, and H.~Fu, ``Mm-retinal: Knowledge-enhanced foundational pretraining with fundus image-text expertise,'' in \emph{International Conference on Medical Image Computing and Computer-Assisted Intervention}, 2024, pp. 722--732.

\bibitem{jing2018automatic}
B.~Jing, P.~Xie, and E.~Xing, ``On the automatic generation of medical imaging reports,'' in \emph{Proceedings of the 56th Annual Meeting of the Association for Computational Linguistics (Volume 1: Long Papers)}, 2018, pp. 2577--2586.

\bibitem{pelka2018roco}
O.~Pelka, ``Radiology objects in context (roco): A multimodal image dataset,'' in \emph{Intravascular Imaging and Computer Assisted Stenting and Large-Scale Annotation of Biomedical Data and Expert Label Synthesis}, D.~Stoyanov, Ed.\hskip 1em plus 0.5em minus 0.4em\relax Springer International Publishing, 2018, pp. 180--189.

\bibitem{papineni2002bleu}
K.~Papineni, S.~Roukos, T.~Ward, and W.-J. Zhu, ``Bleu: a method for automatic evaluation of machine translation,'' in \emph{Proceedings of the 40th Annual Meeting of the Association for Computational Linguistics}, 2002, pp. 311--318.

\bibitem{banerjee2005meteor}
S.~Banerjee and A.~Lavie, ``Meteor: An automatic metric for mt evaluation with improved correlation with human judgments,'' in \emph{Proceedings of the Acl Workshop on Intrinsic and Extrinsic Evaluation Measures for Machine Translation and/or Summarization}, 2005, pp. 65--72.

\bibitem{lin2004rouge}
C.-Y. Lin, ``Rouge: A package for automatic evaluation of summaries,'' in \emph{Text Summarization Branches Out}, 2004, pp. 74--81.

\bibitem{zhao2024ratescore}
W.~Zhao, C.~Wu, X.~Zhang, Y.~Zhang, Y.~Wang, and W.~Xie, ``Ratescore: A metric for radiology report generation,'' in \emph{Proceedings of the 2024 Conference on Empirical Methods in Natural Language Processing}, 2024, pp. 15\,004--15\,019.

\bibitem{tan2024medical}
Y.~Tan, C.~Li, J.~Qin, Y.~Xue, and X.~Xiang, ``Medical image description based on multimodal auxiliary signals and transformer,'' \emph{International Journal of Intelligent Systems}, vol. 2024, no.~1, p. 6680546, 2024.

\bibitem{he2016deep}
K.~He, X.~Zhang, S.~Ren, and J.~Sun, ``Deep residual learning for image recognition,'' in \emph{Proceedings of the IEEE Conference on Computer vision and Pattern Recognition}, 2016, pp. 770--778.

\bibitem{deng2009imagenet}
J.~Deng, W.~Dong, R.~Socher, L.-J. Li, K.~Li, and L.~Fei-Fei, ``Imagenet: A large-scale hierarchical image database,'' in \emph{2009 IEEE Conference on Computer Vision and Pattern Recognition}, 2009, pp. 248--255.

\bibitem{loshchilov2018decoupled}
I.~Loshchilov and F.~Hutter, ``Decoupled weight decay regularization,'' in \emph{International Conference on Learning Representations}, 2019.

\bibitem{vinyals2015show}
O.~Vinyals, A.~Toshev, S.~Bengio, and D.~Erhan, ``Show and tell: A neural image caption generator,'' in \emph{Proceedings of the IEEE Conference on Computer Vision and Pattern Recognition}, 2015, pp. 3156--3164.

\bibitem{xu2015show}
K.~Xu, J.~Ba, R.~Kiros, K.~Cho, A.~Courville, R.~Salakhudinov, R.~Zemel, and Y.~Bengio, ``Show, attend and tell: Neural image caption generation with visual attention,'' in \emph{International Conference on Machine Learning}, 2015, pp. 2048--2057.

\bibitem{lu2017knowing}
J.~Lu, C.~Xiong, D.~Parikh, and R.~Socher, ``Knowing when to look: Adaptive attention via a visual sentinel for image captioning,'' in \emph{Proceedings of the IEEE Conference on Computer Vision and Pattern Recognition}, 2017, pp. 375--383.

\bibitem{yang2021automatic}
S.~Yang, J.~Niu, J.~Wu, Y.~Wang, X.~Liu, and Q.~Li, ``Automatic ultrasound image report generation with adaptive multimodal attention mechanism,'' \emph{Neurocomputing}, vol. 427, pp. 40--49, 2021.

\end{thebibliography}

\end{document}